# Why is That a Good or Not a Good Frying Pan? - Knowledge Representation for Functions of Objects and Tools for Design Understanding, Improvement, and Generation

**Seng-Beng Ho**[1]

[1]Institute of High Performance Computing, A*STAR, Singapore
hosb@ihpc.a-star.edu.sg, hosengbeng@gmail.com

## Abstract

The understanding of the functional aspects of objects and tools is of paramount importance in supporting an intelligent system in navigating around in the environment and interacting with various objects, structures, and systems, to help fulfil its goals. A detailed understanding of functionalities can also lead to design improvements and novel designs that would enhance the operations of AI and robotic systems on the one hand, and human lives on the other. This paper demonstrates how a particular object – in this case, a frying pan – and its participation in the processes it is designed to support – in this case, the frying process – can be represented in a general function representational language and framework, that can be used to flesh out the processes and functionalities involved, leading to a deep conceptual understanding with explainability of functionalities that allows the system to answer "why" questions – why is something a good frying pan, say, or why a certain part on the frying pan is designed in a certain way? Or, why is something *not* a good frying pan? This supports the re-design and improvement on design of objects, artifacts, and tools, as well as the potential for generating novel designs that are functionally accurate, usable, and satisfactory.

## 1 Introduction

While an intelligent autonomous system (IAS) navigates around and interacts with the environment and the objects within it, there are at least two aspects of intelligent processing that feature prominently in its activities. Firstly, it has to recognize the structures and objects in the environment, and secondly, it has to understand what they *afford* it – i.e., the functions of these structures and objects that may support its activities and problem solving processes. Currently, the field of computer vision concerns primarily the former – i.e., object or structure recognition: labeling an object (e.g., a cup) or structure (e.g., a wall) as belonging to a known class (Bishop 2006). Much less work in computer vision in particular and in AI in general has been devoted to the second aspect – function understanding (Dimanzo et al. 1989; Freeman and Newell 1971; Gibson 1979; Grabner et al. 2011; Ho 1987; Stark and Bowyer 1991; Wu, Misra, and Chirikjian 2020; Yao, Ma, and Fei-Fei 2013). This paper proposes a general knowledge representation language and framework that allows an IAS to characterize and understand the functions of objects, structures, and systems, hence supporting its intelligent activities such as problem solving processes to satisfy its goals and needs.

There have been some attempts at recognizing the functionalities of objects and tools (Grabner et al. 2011; Ho 1987; Stark and Bowyer 1991; Wu et al. 2020; Yao et al. 2013). However, among these research works, there are varying degrees of the depth of understanding involved. For example, the effort in (Yao et al. 2013) is merely attempting to distinguish the functionalities of various musical instruments through the recognition of the poses of humans interacting with them, without really characterizing exactly *how* the constructions of various instruments lead to certain sounds and certain ways a musician would interact with it. There is also the question of *why* the instrument involved is designed in a certain way. In Ho (Ho 1987) and Wu (Wu et al. 2020), the detailed constructions of the objects involved are represented, and interactions between objects and humans are considered and simulated to elicit the functionalities involved, going one step further than (Yao et al. 2013) in characterizing the understanding of the functions. However, there is still no deeper level of understanding such as being able to explain *why* certain constructions are present in the object/tool, and there is also no general framework proposed to unify all kinds of situations involving functioning and functions of objects/artifacts/tools.

Ho (Ho 2022) laid out a unified and general framework for characterizing functions, using a representational language that employs a fixed, limited, and general set of concepts and links for representing functional concepts in general situations (the CD+ representational framework). In this paper, this framework is applied to answer specific "why" questions concerning the functionalities of a cooking tool – the frying pan. This work focuses on taking just this one tool to explore the depths of the processes and representations involved – i.e., from the issues of representing and characterizing human activities to the elicitation and representation of certain functionalities involved.

Currently there are generative AI systems that can generate many novel variations of objects, structures, and designs (Deitke et al. 2022; Kolve et al. 2022). Our proposed method would take these generated forms as starting point and evaluate the suitability of these novel forms for certain functionalities. As a result, the design of objects could be improved accordingly, and novel objects could be identified as well.

This work does not provide detailed computer implementations including simulations of the processes involved, but it focuses on laying out the entire process of representing and reasoning with functional processes and concepts, which could be implemented in future work.

## 2 Motivation and Background

Unllike the process of recognition, where the end result is usually a label attached to a certain structure or object, the end result of the process of function understanding involves some descriptions of the function and functioning of the system, structure, and object involved. For example, to characterize the *function* of a kettle, we might use a natural language description such as "a kettle is used to boil water." To characterize the *functioning* of a kettle, we might say, "a kettle works by employing some electrical heating elements placed in its interior, which, when electricity is passed through them, they heat up, and the heat is transferred to the water that is also placed in the kettle's interior in contact with these heating elements. After some time of heating the water, it reaches the boiling point and boils."

Now, the above description of the function and functioning of a kettle in the form of natural language is easily understood by an average human being. But, in order for a machine or IAS to really understand the function and functioning involved, the "meaning" or the deep structures of the sentences above have to be encoded, elucidated, and represented in some machine processable computational structures. Meaning representation is usually addressed by the fields of linguistic semantics (Evans and Green 2006; Geeraerts 2010; Langacker 2008; Talmy 2000) and computational linguistics (Clark, Fox, and Lappin 2012; van Eijck and Unger 2010; Mitkov 2005).

Among the various attempts in linguistics and computational linguistics, it has been identified in (Ho 2022) that conceptual dependency (CD) theory by Schank (Schank 1973, 1975; Schank and Abelson 1977; Schank and Rieger III 1973) is the most suitable deep meaning representational framework to be applied to the characterization of functionality. The main reason for this is that an adequate functional characterization of any system, structure, or object calls for a representational systems that can deal with specifications of complex *causalities* between events, objects, and structures, and CD provides just such a causal framework. Ho (Ho 2022) extended the original CD framework to CD+, incorporating new constructs that provide for a more complete framework for the representation of functionality.

In the earlier work on functionality mentioned in the Introduction, including work on qualitative physics that dealt with the functioning of systems (Bobrow 1985; Faltings and Struss 1992; Forbus 2018; Hobbs and Moore 1985; Weld and de Kleer 1990), there has been no attempt to elucidate a *general* language and framework that supports machine understanding. The CD+ framework is general and is fleshed out in detail in this paper to represent the functional concepts of certain artifact/tool.

Another relevant domain of AI research is the recent work on recognizing and characterizing human activities in various environments (Fu et al. 2022; Gan et al. 2021; Gao et al. 2019; Inamura and Mizuchi 2021; Li et al. 2021; Puig et al. 2018; Xiang et al. 2020; Yang et al. 2015). In these works, there are occasional uses of symbolic descriptions of the activities involved. However, the symbolic constructs used are relatively simple and are not adequate for capturing the complex causal structures inherent in the characterizations of many complex functional concepts, as we shall see in this paper where we instead use CD+ to successfully achieve this end.

## 3 The Representational Framework

It is instructive that we first describe the basic representational constructs of CD+ (Ho 2022).

Figure 1 provides three examples of CD+ representations. In Figure 1(a), the event represented is "Person pushes the door open." In CD+, an event is a "conceptualization" linking a subject and an action. In this event, "Person" is the subject, and "PUSH" is the action. The link linking them is a "conceptualization" link (a double arrow). The object of PUSH in this case is Door - an object ("o") link is used. There is actually a causality present in this statement that is not explicit, that is, the pushing of the door *causes* the door to acquire the *state* of "Open." A causal link is a downward arrow with a vertical line going down its center. The double arrow with a line going down its center is the "state" link.

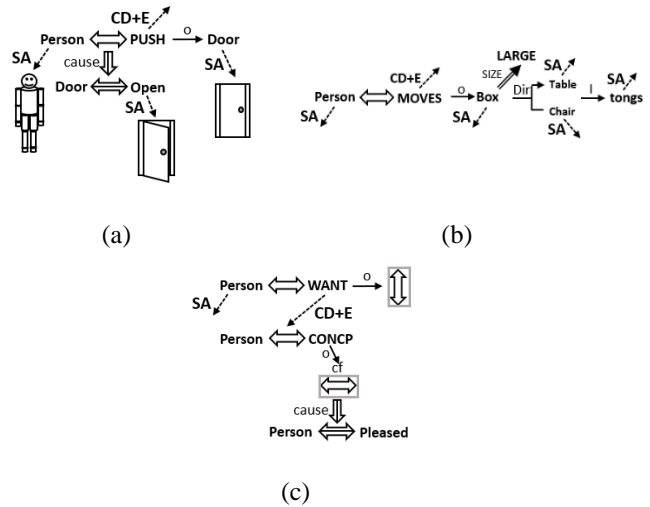

Figure 1. CD+ representations for illustrating the basic CD+ constructs. (a) "Person pushes the door open." (b) "Person moves the box from the chair to the table with a pair of tongs. (c) Person wants something.

In addition, there are two kinds of constructs that elaborate on the symbols in this representation. The Structure Anchor (SA) is a structural specification of symbols that are objects. Person and Door are both symbols pointing to certain real world physical constructs through SA. These structures are detailed models that could be represented by using, say, vectoral constructs much like how objects are represented in a graphics rendering system. Another kind of elaboration is called the CD+ Elaboration (CD+E). This kind of link elaborates actions, and points to a specification of how the action involved is executed. The elaboration could in turn be specified in the form of a CD+ representation. For example, the PUSH action could consist of a number of steps such as "Touch the Surface of the Door with your hand," "Exert a force away from self," etc. CD+Es could be recursively extended down several levels, to a "ground" level such as in the forms of procedural movement routines used in robotics.

Figure 1(b) is a representation of the event "Person moves the large box from the chair to the table." The Dir link specifies the direction of movement. The modifier of the box, LARGE, can be specified using a "property" link, in this case a SIZE property. Suppose an instrument is used, such as "Person moves the box from the chair to the table using a pair of tongs," it can be specified in an Instrument (I) link as shown. The precise way that the movement is executed is specified in the CD+E attached to MOVE. The instrument used in executing the movement could also be specified as part of this CD+E, hence the "I" link may not always be necessary.

Figure 1(c) is the representation of the conceptualization "Person wants something." When a person wants something, the person CONCePtualizes (CONCP) that if that something, which could be an event, represented as another conceptualization (shown as the usual double arrow bounded by a box), were to come true in the future, the person is "Pleased." In the figure, "c" stands for "if" (conditional), and "f" stands for "future."

Having described the basic representational constructs, we next turn to the fleshing out of the function of a frying pan.

## 4 The Representation of Frying and the Function of a Frying Pan

In this section, we describe the processes of cooking and frying using CD+.

Given any activity, whether it is those of more immediate current concern such as baking, frying, cooking, or all other activities that are associated with human day-to-day functioning, there is usually background knowledge associated with them. For an IAS or AI system, this kind of knowledge could be built-in manually by the system creator, crowd-sourced from the Internet, or automatically learned, extracted, generalized, and encoded from prior observations and experiences (Puig et al. 2018). The background knowledge "sometimes people like to eat cooked food" puts cooking in the context of certain human needs. We begin with a discussion on this.

### 4.1 Basic Functioning and Process of Cooking and Eating – Background Knowledge

As mentioned above, to understand and represent the functionality of any particular tool or artifact, often it is necessary to broaden the view by looking at the context in which is it used. A frying pan is part of the cooking and eating process. We begin with representing certain basic requirements and rules often associated with cooking and eating.

Figure 2(a) shows the representation for the conceptualization of "Person wants to eat well-cooked food." (Of course, a person may also want to eat raw food at some other times.) This is the same representational structure as that of Figure 1(c). The state of "well-cookedness" is represented as COOKED(WELL). Note that the state of "well-cookedness" (W) determines in a fuzzy way, Person's satisfaction (S). This in turn determines the degree of being "Pleased" (P).

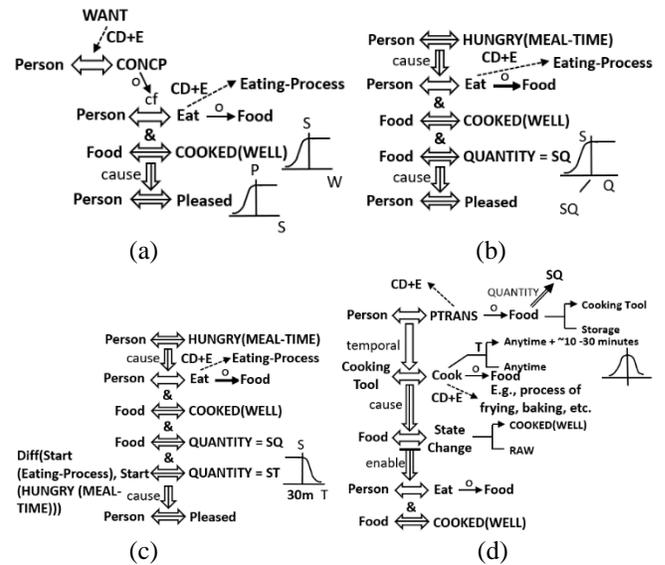

Figure 2. Background knowledge on cooking and eating. (a) Person wants to eat well-cooked food. (b) Person wants to eat a Satisfactory Quantity (SQ) of well-cooked food when hungry at mealtime. (c) When hungry at mealtime, Person wants to eat a Satisfactory Quantity (SQ) of food, and Person also wants to start eating within a Satisfactory amount of Time (ST, typically about 30 minutes) of feeling hungry. (d) A typical cooking process. Here we introduce an "enable" link – the *cause* arrow topped with a horizontal line (Ho 2022).

In a number of places in this paper we use predicate-like representations such as COOKED(WELL), In(Spatula, Frying-Pan), Centered-On(Frying-Pan, Burner), etc. and we often omit the first argument if it is clear, such as Centered-On(Burner) in Figure 3(a) because this is part of the elaboration of the main concept Frying-Pan.

An interesting question is, how is the state of something being cooked or well-cooked defined, determined, and recognized? From the point of a human being, she has internal sensing criteria in her gustatory system to define what is meant by something being "cooked" or "cooked-well," Tasting something raw vs tasting something cooked are certainly very distinguishable gustatory experiences. Humans also often experience and learn the association of certain visually identifiable attributes (externally or internally) and the degree of being cooked. (Note that a piece of food may look cooked on the outside but not cooked inside, but humans can rely on either what is visible from the outside to infer what is happening inside (e.g., if the outside looks "very cooked," the inside may be just cooked), or rely on other parameters from her experiences – e.g., if the outside of a piece of food of a certain shape and size has started to look cooked, it may take a certain further amount of time for the inside to be cooked as well.) Thus COOKED(WELL) can be determined by a human or AI through vision.

Figure 2(b) shows the representation for the conceptualization of "Person wants to eat a Satisfactory Quantity (SQ) of well-cooked food when hungry at mealtime." Given a certain quantity of food (Q), the degree of satisfaction (S) is a fuzzy function of Q, and HUNGRY(MEAL-TIME) is also a fuzzy function – above a certain degree of hunger, the *want* to eat food is (strongly) triggered (the *cause* link on top). Pleased is now a function of COOKED(WELL) and SQ.

There could be situations in which a smaller quantity of food is first cooked and served, and then as the person is eating, more is cooked and served (e.g., the Japanese dining style known as Omakase). But the above is *one* satisfactory situation.

Figure 2(c) shows the representation for the conceptualization of "When hungry at mealtime, Person wants to eat a Satisfactory Quantity (SQ) of food, and Person also wants to start eating within a Satisfactory amount of Time (ST, typically about 30 minutes) of feeling hungry."

In Figure 2(d) we show the representation for a typical cooking process. The first step of the process is Physically TRANSfering (PTRANS) the food to be cooked from the Storage area to the Cooking Tool (Frying pan for frying, oven for baking, etc.). A new kind of link, "temporal," is introduced here in which the first conceptualization flows to the next one after some time. The next conceptualization involves using the Cooking Tool to Cook Food. This causes a change in the state of Food from RAW to COOKED(WELL). And in turn this enables Person to Eat the well-cooked food, which leads to Person being Pleased if combined with Figure 2(a).

How are the various causal links in Figures 2(a) – (d) learned? Consider Figure 2(a). We have mentioned above that the internal pleasure a person feels when she consumes well-cooked food is a built-in need, arising from certain biological properties within the human body (the gustatory responses, etc.). Through a causal learning process which is not explored in this paper but which has been extensively studied elsewhere (Fire and Zhu 2013, 2016; Ho, Edmonds, and Zhu 2020; Ho and Liausvia 2016; Ho, Yang, and Quieta 2020; Yang and Ho 2018), this *temporal* association (first, put well-cooked food in mouth, then experience pleasure) is identified as a *causal* relation. (Another reasoning process that could lead to this causal establishment is a counter-factual process: had the food not been well-cooked and I put it in my mouth, I would not be Pleased. Therefore, it is the well-cookedness of the food that causes me to be Pleased.)

The other causal links in Figure 2(d) can be learned likewise.

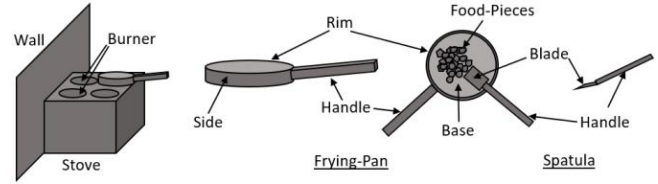

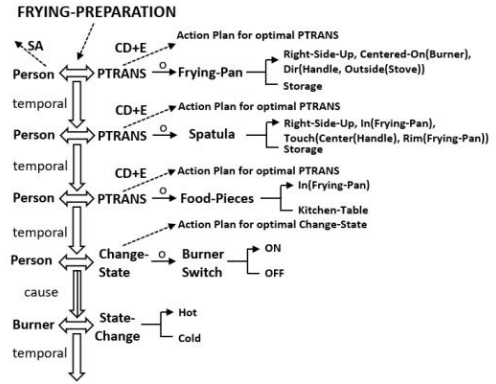

(a)

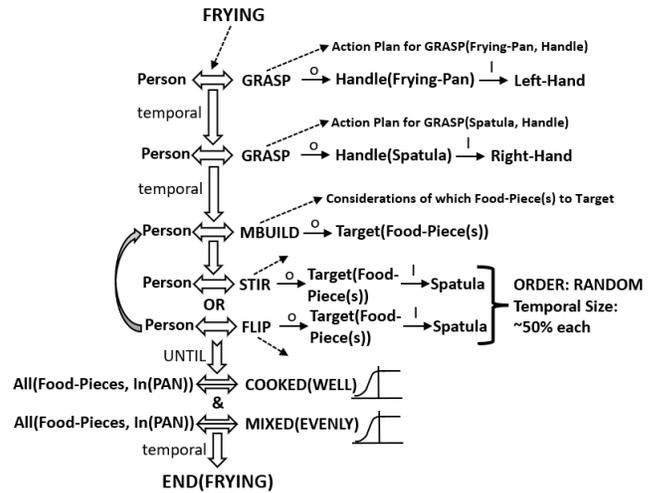

(b)

Figure 3. (a) FRYING PREPARATION process. (b) FRYING process.

## 4.2 Observed Functioning and Process of Frying

Assume that there is a process similar to that described in prior work such as that in (Gao et al. 2019; Puig et al. 2018; Yang et al. 2015), in which human daily activities in a house, say, are observed, learned, and encoded. Further assume that the details of each hand movement, and movements of the objects involved that the human hands interact with, are observed and encoded in CD+ (granted that this is not a trivial process). A possible CD+ encoding of the frying process is shown in Figure 3.

In Figure 3, we divide the process into FRYING-PREPARATION and FRYING proper (in reality these stages are contiguous). Figure 3(a) shows that the FRYING PREPARATION process consists of first PTRANSing the Frying-Pan from the Storage to the Burner on a Stove. It is positioned Right-Side-Up, Centered-On the Burner, and is oriented with the Handle facing the Outside of the Stove. The meaning of these spatial terms are pre-defined as grounded concepts or near-ground concepts as articulated in (Ho 2022; Ho and Wang 2019). These concepts are quite general, and once defined, can be applied to a large number of situations involving physical objects of all kinds.

In this first step, the details of the PTRANS action is described in a CD+E which is the "Action Plan for (Optimal) PTRANSing of Frying-Pan from Storage to the top of the Burner, positioned accordingly." This Action Plan can be learned from observation, or obtained from a motion planning process. For example, very sophisticated motion planning and collision detection processes such as described in (Ruan et al. 2023; Ruan, Wang, and Chirikjian 2022) that can avoid arbitrarily narrow obstacle spaces can be used.

The other steps are self-explanatory, and the process ends with Burner changing state from Cold to Hot.

After the process of FRYING-PREPARATION, the process of FRYING begins. Typically, for a right-hander, Person would use Left-Hand to GRASP Handle(Frying-Pan) and Right-Hand to GRASP Handle(Spatula).

Also, typically, when a person is frying something, she would sometimes *stir* the food pieces around and sometimes *flip* the food pieces so that different sides of the food pieces get cooked. These two actions may take place in any sequence, but they each may be executed, say, roughly equally often (real-world statistics gathering for individual frying sessions may reveal other kinds of distributions). At the beginning of each action, the person usually forms some thoughts about which food-piece should be "operated on" next (e.g., "that piece looks uncooked – e.g., the top surface exhibits visual signs of being uncooked - let's target to flip it next…", or "these pieces are not well mixed, let's target to stir them around, " etc.). This is an MBUILD (Mental reasoning to BUILD some internal mental representations) process in CD+ (Ho 2022; Schank 1973).

Figure 3(b) shows that at this stage of the frying process, for each step, Person first MBUILDs the Target(Food-Pieces) Then Person either STIR or FLIP the targeted Food-Pieces(s) (whether to STIR or FLIP could be part of the MBUILD process). These actions are repeated until certain conditions are met.

A typical end-process test would be when "all food pieces in the frying pan are cooked AND all food pieces in the frying pan are evenly (well) mixed." Mixing or mixing evenly becomes relevant when there is more than one kind of food pieces to be fried. Having the food well mixed is a basic "want" of a typical person.

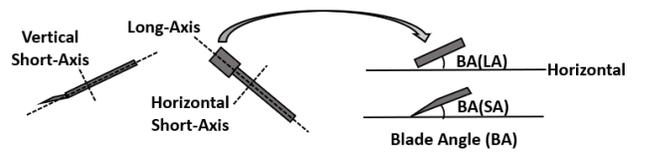

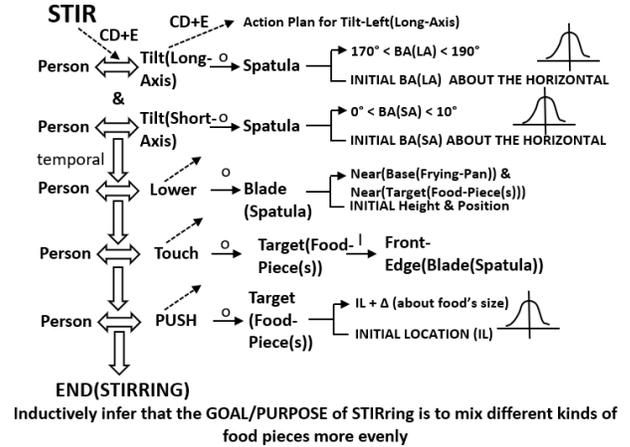

(a)

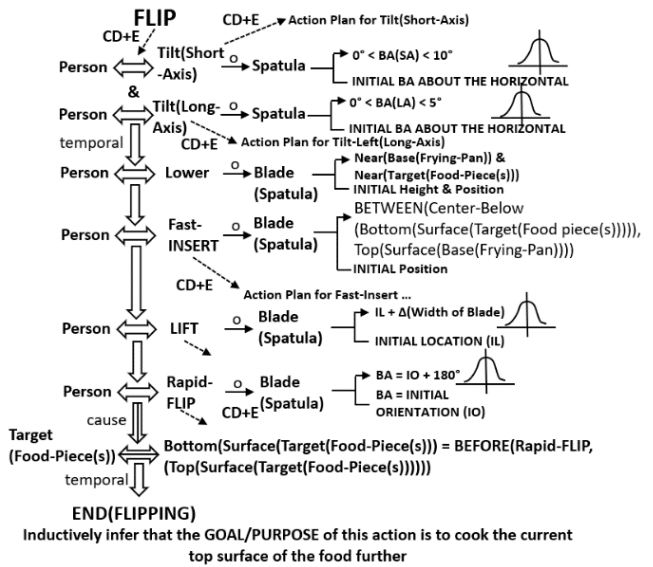

(b)

Figure 4. (a) STIRring process. (b) FLIPping process.

The criteria of "COOKED(WELL)" and "MIXED(EVENLY)" are also fuzzy functions. The definition of "MIXED(EVENLY) is described in Section 4.7.

Now, for an AI system to understand that these are the terminating conditions of the stirring and flipping processes, it could inductively derive these through many instances of observation, or through background knowledge such as that in Figure 2(a).

Figure 4 shows the processes of STIRring and FLIPping. These are self-explanatory. Longish objects have a long axis and two short axes (Ho and Wang 2019). (Figure 7(a) illustrates the positioning of the spatula in the STIRring mode.)

One thing to note is that in the FLIPping process, the spatula has to be inserted underneath the food pieces in a "fast" insertion process, otherwise, Blade(Spatula) would just push the target food-pieces around and not be positioned for the FLIPping step. Also, after LIFTing the food, a Rapid-FLIPping action is executed, otherwise, instead of flipping the food-piece(s), they may just fall back to the frying pan with the respective faces still facing in the same directions. We made assumptions on the various angles indicated n Figure 4 but they are supposed to be derived from real-world observations.

There could yet be other ways to STIR and FLIP Food-Piece(s) and these could be added to the processes in Figure 4 as disjunctive processes to achieve the same ends.

### 4.3 Cooking/Frying Success Conditions

There are two further conditions that must be tested before the success of the cooking/frying process could be ascertained.

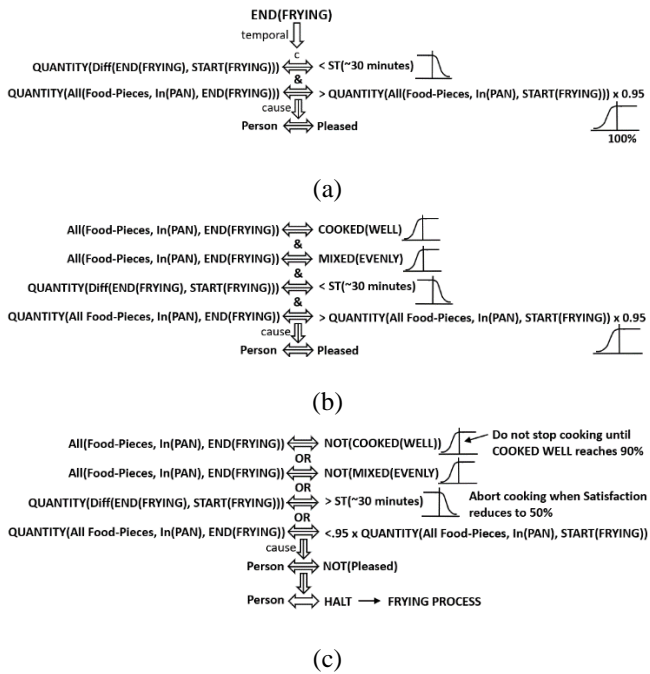

Figure 5. (a) Satisfaction from the time taken to cook and the amount of food that is retained in the frying pan. (b) Conditions in (a) plus the conditions in Figure 3(b). (c) The contrapositive of (b).

As stated in Figure 2(c), a person typically wants to start eating a sufficient quantity (SQ) of food within a certain time (ST) of beginning to feel hungry. Likewise, if observations are made on people frying, baking, or use whatever method to cook food, each process also typically would not take too long a time. Let's assume that this time is also about ST and it is about 30 minutes. Then, inductively, the system could infer that completing each frying or whatever kind of cooking process within ST is a condition of "satisfaction," and that Person with be Pleased. This is stated in Figure 5(a).

Now, in whatever cooking process, especially in a frying process, food pieces could be lost due to excessive STIRring or LIFTing and FLIPping. Hence, another satisfaction condition is that the quantity of food pieces at the end of the frying process should be, perhaps, typically larger than 95% (say, through observation what is typically acceptable) of what the frying process started with. The satisfaction with regard to this is also a fuzzy function.

Figure 5(b) combines all the 4 conditions, including COOKED(WELL) and MIXED(EVENLY).

Figure 5(c) is a contrapositive of Figure 5(b), which is that if *any* of the 4 conditions is not met, Person would not be Pleased and it would cause Person to HALT the process.

The 4 conditions of Figure 5(c) could be activated in multiple ways. It could be that there was an attempt to finish the cooking process as stipulated in Figure 3(b) (COOKED(WELL) and MIXED(EVENLY)), and then the ST or the 95% condition is violated, and the process comes to a halt, or it could be that the cooking process of Figure 3(b) stops when ST is exceeded or when the 95% rule is exceeded (Figure 3(b)'s terminating conditions would have to be modified to represent this), and it is found that COOKED(WELL) or MIXED(EVENLY) is not satisfied.

Later in Section 4.4, Figure 5(c) would be used to judge if certain frying tools (Frying-Pan and Spatula) with certain designs can adequately function to support frying.

### 4.4 Understanding the Frying Process

Up to this point, what has been encoded and represented is a description of the causal and temporal processes of a particular kind of cooking activity – frying (as can be seen in Figures 1-6, the "backbones" of the representations involved are composed of either a *causa*l link or a *temporal* link). These representations reflect certain levels of understanding of the processes involved. For example, the temporal links can be used to answer "what" questions. If the question is "What happens after Food-Pieces are placed inside Frying-Pan?" The answer would be "Person then turns on Burner." (Figure 3(a)). Causal links could be used to answer "why" questions. For example, if the question is "Why Person is not Pleased with the frying process?" The answer could be "Because the frying process took more than 30 minutes." (Figure 5(c)). At this stage, the system does not seem to possess the representations for answering questions related to the causal functions of the frying pan and the spatula, and for that matter, the burner, in supporting the cooking process of turning the food pieces from uncooked to well-cooked, and also well-mixed in the frying pan. If we look at Figure 3(a), it does show that Burner is turned on, and it changes from Cold to Hot, but there is no knowledge that it is the heat that is transmitted through the frying pan placed on top of Burner, that in turn gets transmitted to the food placed in the frying pan that causes the food to be cooked.

Similarly, in Figure 3(b), it is the stirring and flipping actions that resulted in the well-mixing and well-cookedness of the food pieces, but there is no explicit causal links here.

There needs to be further background knowledge of physics of heat and heat transfer across metal, as well as physics of the more mechanical aspects of things such as pushing, lifting, flipping, etc. that will elicit the causal understanding.

If the interests of paper length, in the next section we focus on encoding and representing some basic physics knowledge on the pushing and moving of objects to elicit certain functionalities associated with the frying process, thus the functional roles of Frying-Pan and Spatula in the process.

### 4.5 Physics of Pushing and Moving

In Figure 6, we illustrate several rules regarding the pushing and moving of objects. This will later be applied to the understanding of the pushing and moving of food pieces in the frying pan to elicit certain features of the frying pan.

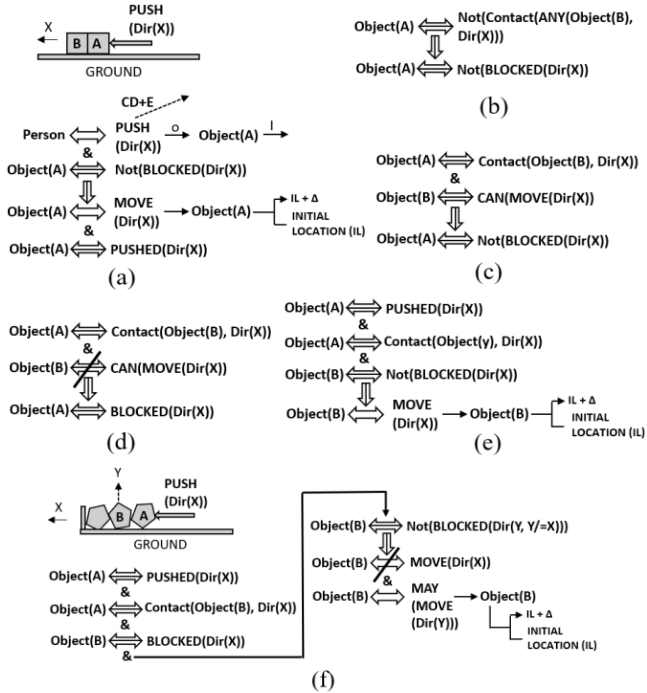

Figure 6. The physics rules governing forces and objects. (a) Basic PUSH action. (b) Not-BLOCKing condition. (c) Transmitted Not-BLOCKing condition. (d) Transmitted BLOCK condition. (e) Transmission of unblocked condition. (f) Movement in an unblocked direction when some other directions are blocked. Note that "Not" can be represented in two ways: as a predicate Not or a slash across the negated conceptualization.

Figure 6(a) is the most basic rule of the pushing process. It states that if Person PUSHes Object(A) in direction Dir(X) with some instrument (could be the hand), and Object(A) is not BLOCKED in Dir(X), then it causes Object(A) to MOVE in Dir(X) by a $\Delta$ amount. (We are assuming this is an incremental act of PUSHing that causes Object(A) to move just a little. Sustained PUSHing will of course lead to continued displacement.) At the same time, Object(A) acquires a state of being PUSHED in Dir(X). Figure 6(b) – (e) are self-explanatory. The operational meaning of CAN (Figure 6(d)) is discussed in (Ho 2022).

Figure 6(f) illustrates a situation in which Object(B) is BLOCKED in Dir(X) but Not BLOCKED in Dir(Y), Y being different from X, then there is a possibility that when Object(A) is PUSHed, Object(B) MAY MOVE in Dir(Y). The operational meaning of "MAY" has also been discussed in (Ho 2022).

These rules of physics are learnable in a symbolic causal learning process akin to what has been discussed in (Fire and Zhu 2013, 2016; Ho, Edmonds, et al. 2020; Ho and Liausvia 2016; Ho, Yang, et al. 2020; Yang and Ho 2018).

### 4.6 The Function of the Frying Pan's Side

In this section we begin with the understanding of how the frying pan functions to support cooking in general and frying in particular, before we present the processes that identifies what is wrong with certain other designs that cannot fulfill the function in a satisfactory manner.

The title of this section seems a little narrow – we are only looking at the function of one part of the frying pan - its side, shown in Figure 3(a). This is mainly because of the small sets of physical knowledge that was encoded and represented in the previous sub-section. For example, to understand the function of the base of the frying pan, which is partly to transmit heat from Burner to Food-Pieces, we need background knowledge on heat. However, the consideration of the function of the frying pan's side provides the paradigm for reasoning about the other parts, or for that matter, other objects.

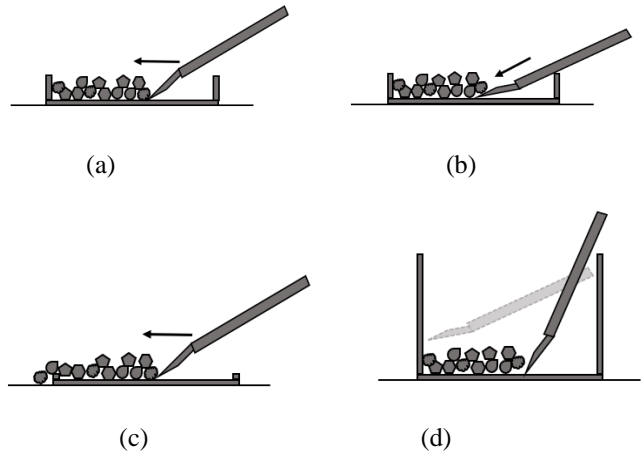

Figure 7. (a) The STIRring situation. (b) The FLIPping situation. (c) A "frying pan" with a very low side. (d) A "frying pan" with an excessively high side. The handle is omitted here.

Now, the set of physical rules in Figure 6 can help an AI system understand the causalities involved and hence the attendant functionalities. The base of the frying pan is now the new "GROUND." The side of the frying pan is fixed to the base and hence is immobile relative to the base. In Figure 7(a) we show the cross-section of the stirring process, which, according to the encoding in Figure 4(a), involves flipping the spatula "upside-down" as shown and the edge of the blade is used to touch and push the food pieces involved.

Now, we can set this Figure 7(a) configuration up in a simulator such as the Unity engine (www.unity.com) and have it simulate what would happen after executing some stirring action. The processes outlined in Figure 3(b) (of the entire frying process) or Figure 4(a) are used to direct the simulation process. Naturally, the simulator would generate the expected consequences: the food pieces staying within the confines of the frying pan.

However, the physics reasoning engine of Unity has all the physics rules built-in and they are hence implicit and cannot be tapped directly for reasoning about causality. This is when the explicitly encoded rules of Figure 6 becomes useful.

According to the physical rules of Figure 6, this pushing action would transmit forces over some of the food pieces, and ultimately the forces would reach some of the food pieces right next to and are in contact with the side. Now, as the side is immobile, the food pieces right next to it and are contacting it would be immobile too and stay within the frying pan. We do not show the details of the inference processes here but the combinations of the rules in Figure 6 allow the following conclusion to be derived: "The side of the frying pan *causes* the food pieces *not* to go beyond the confine of the frying pan."

Ho (Ho 2022) describes a counter-factual process that can also be executed to infer the functions of the various parts of an object. In this case, suppose the AI system takes a mental action of *removing* the side of the frying pan, and then execute a simulation of the stirring process of Figure 4(a), and the Unity engine is then used to simulate the consequences, it would find that the food pieces would spill out of the frying pan's confine. (These processes of mental reasoning and activating simulation, termed "MENTAL EXPERIMENTS," can be represented in CD+ format, as shown in (Ho 2022), and the CD+ representational constructs *direct* the MENTAL EXPERIMENTAL processes.) Therefore, the AI system observes that "had the side of the frying pan not been there, the food would spill outside the frying pan during the stirring process," (represented in CD+) and concludes that "the function of the side of the frying pan is to *enable* the food pieces to stay within the base of the frying pan during the stirring process" (represented in CD+).

Figure 7(b) shows the flipping process' first step, which is to rapidly insert the blade of the spatula under some food pieces, and that is followed by lifting of the food and flipping it around. In this process, some of the other food pieces might also get pushed toward the side of the frying pan, and it is the side that prevents them from spilling outside the frying pan.

## 4.7 Why is that a good or not a good frying pan?

We are now armed with the devices that allow an AI system to reason about whether certain tool/artifact/object can satisfy certain functional requirements, and if not, *why* not (i.e., providing explanations). We will consider the standard as well as alternatives of frying pan-like objects to test their functional capabilities with regards to supporting the frying process.

Figure 8 shows the overall process of functional reasoning. Given an object to be tested for frying function (e.g., the standard Frying-Pan of Figures 3(a), 7(a) or 7(b)), the FUNCTIONAL REASONER then deposits the object involved into a SIMULATION ENVIRONMENT (a buffer of sort). (The structure of Figure 8 had been earlier proposed in (Ho 1987, 2022).) Then it follows the processes dictated by Figure 3(a) for FRYING-PREPARATION (this representation is stored in the CONCEPTUAL CORE). Again, something like the Unity simulation engine is deployed here. There is a module called PHYSICAL REASONER that contains both Unity's implicit knowledge (or for that matter, the knowledge in any simulation engine) as well as the explicitly stated physics knowledge illustrated in Figure 6. All these modules and processes are subparts of a more general Unified General Autonomous and Language Reasoning Architecture (UGALRA) described in (Ho 2022).

Next, the FRYING processes described in Figures 4(b) and 5 are executed accordingly. A slightly tricky situation is, a usual simulator may not be able to simulate how well-cooked certain piece of food may look after certain amount of heat is added to it. For just this purpose, we could build in a rule that says that each piece of food has to have each of its various faces touching Top(Surface(Base)) of Frying-Pan for X minutes in total. In a typical, say, 20-30-minute frying process, X could be in the order of 3-5 minutes.

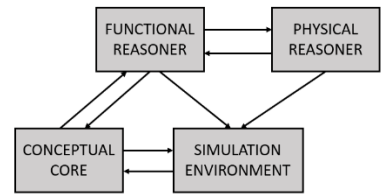

Figure 8. The various modules of FUNCTIONAL REASONER, PHYSICAL REASONER, SIMULATION ENVIRONMENT, and CONCEPTUAL CORE interacting with each other for functional reasoning. See text for explanations.

The MIXED(EVENLY) condition can be defined as follows: the entire area of Food-Pieces is first divided into, say, 10 or more subareas, and if the number of different kinds of food pieces in each of the subareas are all about equal, then the MIXED(EVENLY) condition is met.

The FUNCTIONAL REASONER could first end the FRYING process through the two conditions stated in Figure 3(b) – COOKED(WELL) and MIXED(EVENLY), and then check if the other two conditions, "< ST" and ">95%", stated in Figure 5(a) are met. If so, "Person is Pleased" and the frying pan fulfils its function.

At this point, the system can assume that all the individual steps of the entire frying process must have been carried out correctly, thereby leading to the final positive result. There are two aspects involved: the activities carried out at each step, and the tools involved in supporting those activities. Therefore, Burner must have fulfilled its function, so did Frying-Pan and Spatula. At this stage, the system can respond with "it is a good Burner," "it is a good Frying-Pan," "it is a good Spatula," "it is a good Stove," etc. ("etc." because the list could go on and include "it is a good heat source supplied to the burner" and so on.)

Suppose the system is asked a further detailed question such as "WHY is that a good frying pan?" The system would consider the 4 conditions in Figure 5(b) each in turn to see how the frying pan and its construction cause the condition to be satisfied.

Suppose the system considers the condition "leftover quantity of food pieces in frying pan must be > 95% of what was there in the beginning of the frying process," and tries to identify how the frying pan may cause that to happen, it could carry out a MENTAL EXPERIMENT like mentioned in the previous section, in which the functioning of the frying pan in the process is dissected. At this stage, the system may not know that it is the frying pan and its construction that allow this to happen, and it has to consider all the tools involved and their parts.

Because the system removes ONE factor in each step of its consideration, it can conclude that the observed unsatisfactory consequent is *cause* by THAT factor. Therefore, when the Side of the pan is being considered, the reasoning is, "had Frying-Pan not have Side, Food-Pieces will not stay in Frying-Pan and the >95% condition fails."

The system could respond with an output such as "That is a good frying pan because its side prevents food pieces from spilling outside of the frying pan in the frying process."

Likewise, there are other aspects of the frying pan's function that can be deduced: "That is a good frying pan because its base is not too small to allow stirring and mixing of food evenly," "That is a good frying pan because its base is made of metal which allows heat to transfer from the heat source to the food," etc.

As mentioned before, other than conducting counter-factual reasoning, the physics knowledge in Figure 6 could also support the reasoning that it is the Side of the frying pan that "blocks" the food pieces from falling outside, and answer the WHY question accordingly,

Suppose in another mental experiment, the system removes Blade of Spatula instead (Frying-Pan and other things stay constant), then the consequence would be that the STIRring process is very inefficient and the FLIPping process is nearly impossible. Food-Pieces will still get pushed around, resulting in some mixing, and some may occasionally roll over, resulting in all sides of Food-Pieces get cooked, but by the time the process finishes, it would be much longer than ST. In this experiment, the >95% rules may not be violated.

The system can continue to drill down further to understand WHY a Spatula without Blade or with a poorly designed Blade cannot function well to support the process of frying.

In Figure 7(c), we show a frying pan that has a very low side. A similar process of reasoning will lead to the conclusion that "That is NOT a good frying pan," and "That is not a good frying BECAUSE its side is too low and in the process of frying food-pieces will spill outside the pan and much more than 95% of food-pieces will be lost."

Figure 7(d) shows a situation in which the side of the frying pan is way too high. In simulation, it is likely that the time taken for "COOKED(WELL)" will be too long because the spatula cannot effectively flip the food pieces around so that all the sides of the food pieces can be properly cooked within a reasonable time (<ST), though in principle the sides could still ultimately be cooked as the pushing action may still flip the pieces occasionally. Figure 7(d) also shows that even if the entire Spatula is lowered into Frying-Pan to obtain a good angle for the Blade, the FLIPping action still cannot be properly executed. The MIXED(EVENLY) condition could probably still be met within ST.

In a similar reasoning process as discussed above, it is identified that it is the height of the side of the frying pan that causes the inability in positioning Blade(Spatula) in a correct way as stipulated in Figure 4(b) for the effective executing of the Fast-INSERT step, as a result the FLIPping act cannot be successfully carried out (i.e., no Action Plan exists for FLIPing to be executed properly). Thus, this is WHY the frying pan is not a good frying pan.

A verbal explanatory output of the system could be something like "This frying pan is not a good frying pan because its side is too high, which prevents the spatula from being able to adopt an angle to be able to flip the food pieces, and this leads to a much longer time than that of ST to reached the well-cooked states of all the food pieces involved, and this in turn leads to the person involved not being pleased."

With other knowledge such as heating and burning effects on various materials and the human body, we can present things like a frying pan or a spatula made of paper or porcelain, or with inadequate handle length, etc., and have the system explain why they are not good frying pans, or spatulas. (If the handle of the frying pan or spatula is too short, the human or robot's hand would be burned in the process of frying – i.e., in the process of carrying out the frying actions as stipulated in the representation of Figure 3(b).)

## 5 Conclusions and Future Work

This paper demonstrates that the representation and understanding of functionalities involve several aspects of events taking place in the environment and requires the use of a representation language. These include the relevant contexts in which certain processes and functioning take place (in our particular example, it would be cooking and frying), and in which certain objects and tools participate in the causal processes that support the functionalities involved. There is also usually some background knowledge involved (in our particular case, this would be the eating and cooking background knowledge of Figure 2, and the physics knowledge of Figure 6). And we have successfully demonstrated that whether it is background knowledge, knowledge about the particular causal effects that a certain object or tool have in its interacting with other objects to elicit certain functionalities, or the knowledge about the entire context in which the functioning takes place, in which the adequacy of the objects and tools involved in supporting the desired functionalities is evaluated and understood, they can all be represented using the CD+ representational language and framework (Ho 2022), which is general and consists of a limited number of primitive structures and operations, as well as a limited number of basic grounded concepts (Ho 2022).

This proposed method and framework could work with certain current generative AI systems (Deitke et al. 2022; Kolve et al. 2022) in evaluating the novel designs they generate, thus identifying which novel design would work well functionally in certain given environments to serve certain purposes.

Future work includes investigating how this methodology could be extended to many other daily-encountered tools and objects, as well as more specialized tools and objects found in factories and other business environments such as offices, restaurants, gas stations, etc., to represent and understand the functionalities involved so as to be able to judge how well they can fulfil certain desired functionalities as well as how new and better designs can be concocted. Future work will also flesh out the detailed mental experimental processes in CD+ form for reasoning about design improvement and novel design generation. The other main work to devote future effort to would be a complete program implementation of the representational constructs and the attendant reasoning processes.